\documentclass[sigconf]{acmart}
\usepackage{graphicx}
\usepackage{float}
\usepackage{subfig}
\usepackage{array}
\usepackage{multirow}
\usepackage{makecell}
\usepackage{amsmath}
\usepackage{xcolor}

\copyrightyear{2025}
\acmYear{2025}
\setcopyright{acmlicensed}
\acmConference[KDD '25] {Proceedings of the 31st ACM SIGKDD Conference on Knowledge Discovery and Data Mining V.1}{August 3--7, 2025}{Toronto, ON, Canada.}
\acmBooktitle{Proceedings of the 31st ACM SIGKDD Conference on Knowledge Discovery and Data Mining V.1 (KDD '25), August 3--7, 2025, Toronto, ON, Canada}
\acmISBN{979-8-4007-1245-6/25/08}
\acmDOI{10.1145/3690624.3709195}
\begin{document}
\title{Inductive Link Prediction on N-ary Relational Facts via Semantic Hypergraph Reasoning}

\author{Gongzhu Yin}
\orcid{0000-0002-8251-742X}
\affiliation{%
	\institution{Harbin Institute of Technology}
	\city{Harbin}
	\state{Heilongjiang}
	\country{China}}
\email{yingz@hit.edu.cn}

\author{Hongli Zhang}
\authornote{The corresponding author.}
\orcid{0000-0002-8167-7106}
\affiliation{%
	\institution{Harbin Institute of Technology}
	\city{Harbin}
	\state{Heilongjiang}
	\country{China}}
\email{zhanghongli@hit.edu.cn}

\author{Yuchen Yang}
\orcid{0000-0001-8495-6103}
\affiliation{%
	\institution{Harbin Institute of Technology}
	\city{Harbin}
	\state{Heilongjiang}
	\country{China}}
\email{yangyc@hit.edu.cn}

\author{Yi Luo}
\orcid{0000-0003-1479-6081}
\affiliation{%
	\institution{Harbin Institute of Technology}
	\city{Harbin}
	\state{Heilongjiang}
	\country{China}}
\email{yi_luo@stu.hit.edu.cn}

\begin{abstract}
N-ary relational facts represent semantic correlations among more than two entities. While recent studies have developed link prediction (LP) methods to infer missing relations for knowledge graphs (KGs) containing n-ary relational facts, they are generally limited to transductive settings. Fully inductive settings, where predictions are made on previously unseen entities, remain a significant challenge. As existing methods are mainly entity embedding-based, they struggle to capture entity-independent logical rules. To fill in this gap, we propose an n-ary subgraph reasoning framework for fully inductive link prediction (ILP) on n-ary relational facts. This framework reasons over local subgraphs and has a strong inductive inference ability to capture n-ary patterns. Specifically, we introduce a novel graph structure, the n-ary semantic hypergraph, to facilitate subgraph extraction. Moreover, we develop a subgraph aggregating network, NS-HART, to effectively mine complex semantic correlations within subgraphs. Theoretically, we provide a thorough analysis from the score function optimization perspective to shed light on NS-HART's effectiveness for n-ary ILP tasks. Empirically, we conduct extensive experiments on a series of inductive benchmarks, including transfer reasoning (with and without entity features) and pairwise subgraph reasoning. The results highlight the superiority of the n-ary subgraph reasoning framework and the exceptional inductive ability of NS-HART.
\end{abstract}


\maketitle

\newcommand\kddavailabilityurl{https://doi.org/10.5281/zenodo.14637435}

\ifdefempty{\kddavailabilityurl}{}{
\begingroup\small\noindent\raggedright\textbf{KDD Availability Link:}\\
The source code of this paper has been made publicly available at  \textcolor{blue}{\url{https://github.com/yin-gz/Nary-Inductive-SubGraph}}.
\endgroup
}

\section{Introduction}
\begin{figure}\centering
	\includegraphics[width=0.85\linewidth]{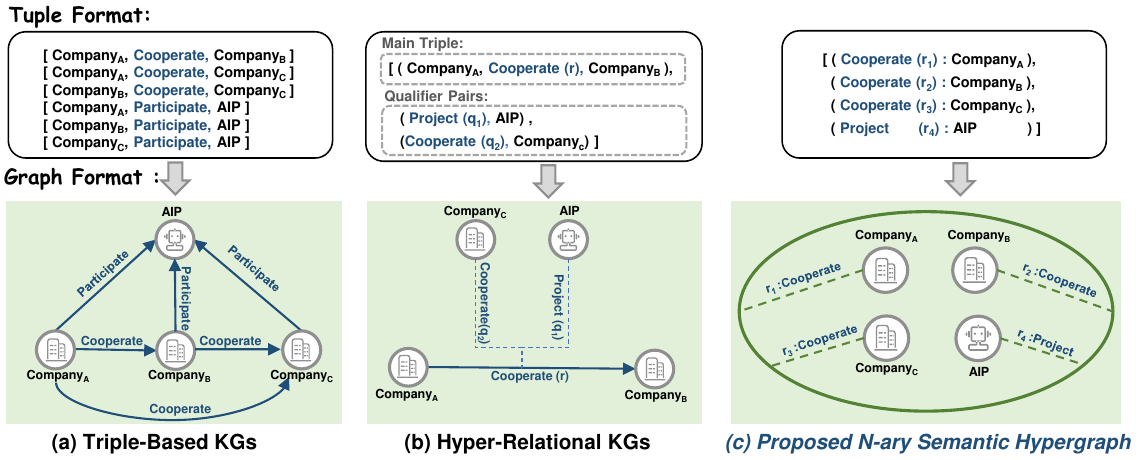}
	\caption{Three ways to represent the n-ary relational fact: "$Company_{A}$, $Company_{B}$, $Company_{C}$ cooperate in project $AIP$." Note that, our proposed n-ary semantic hypergraph is based on the key-value pair representation.}
	\label{fig_example}
\end{figure}

In knowledge graphs (KGs), a "fact" is a piece of information representing semantic correlations between entities. Traditionally, KGs store facts as triples $\left( {h,r,t} \right)$ \citep{ref1}, where $h, t \in \mathcal{E}$ are entities connected by a binary relation $r$. However, recent studies \citep{ref4,ref5,ref6} highlight that triple-based KGs struggle to express n-ary relations, which involve more than two entities and are common in real-world scenarios. For example, in Freebase, over 1/3 of the entities participate in non-binary relations, yet these are decomposed into binary facts, leading to information loss \citep{ref4}. As shown in Figure \ref{fig_example} (a), representing the fact "A, B, and C cooperate on project AIP" using triple fragments the holistic context of "cooperate." For instance, it might incorrectly suggest that A, B, and C independently participate in AIP while cooperating in other projects, failing to capture their joint collaboration on AIP as a unified fact.
	
To tackle this problem, KGs containing n-ary relational facts have emerged \citep{ref4, ref6}. Algebraically, while binary relations are subsets of the Cartesian product $\mathcal{E}^2$, n-ary relations \citep{ref4} extend to the $J$-fold Cartesian product $\mathcal{E}^J$, where $J \geq 2$ is an arbitrary integer. In n-ary relational KGs, n-ary relations carry semantic information by assigning roles to entities, modeling their joint correlations within a fact and avoiding the fragmentation and information loss caused by decomposition. As illustrated in Figure \ref{fig_example} (b) and (c), two prevalent formats exist to represent n-ary relational facts: \textbf{the hyper-relational representation} \citep{ref6} and \textbf{the key-value pair representation} \citep{ref7}. The hyper-relational representation remains fundamentally triple-based, treating a fact as a main triple with additional qualifier pairs, i.e. $[(h, r, t),\left\{\left(q_i: v_i\right)\right\}_{i=1}^{n-2}]$, where $v_i$ is one qualifier entity and $q_i$ is its relation to the main triple. In contrast, the key-value pair representation is inherently n-ary and more flexible, as it uniformly treats a fact as a set of key-value pairs, where each key $r_i$ denotes a role of each entity in this fact, i.e. $[\left\{\left(r_i: v_i\right)\right\}_{i=1}^n]$.

The task of link prediction (LP), which aims at predicting new links for entities based on existing facts, is crucial for completing KGs and inferring potential links\citep{ref2,ref3}. Based on these two formats, several studies \citep{ref6,ref7,ref9,ref10} have developed associated LP methods, focusing on learning low-dimensional embeddings for entities and relations, then making predictions based on these embeddings. For the hyper-relational representation, some studies \citep{ref6} have developed a corresponding graph format by incorporating qualifier information as the attribute of binary edges, thus leveraging graph neighborhood information\citep{ref13, ref14} to enhance embeddings. \textbf{However, for the key-value pair representation, no corresponding graph formats have been provided in existing studies.}

\paragraph{\textbf{Motivation.}}
Most of these LP methods are under transductive settings, assuming all entities are fixed during training and evaluation. This assumption does not align well with the dynamic nature of real-world KGs, which frequently evolve by incorporating entirely new (sub-)graphs, i.e. inductive scenes. One pioneering work \citep{ref16} has initiated exploration into inductive link prediction (ILP) on n-ary relational facts, demonstrating that directly modeling n-ary relations outperforms methods based on triple decomposition. However, its experimental results also revealed that merely extending transductive LP methods yielded suboptimal results in fully inductive scenes (with MRR below 0.1 on most datasets), highlighting the need for further research. 

\begin{figure}\centering
	\includegraphics[width=0.99\linewidth]{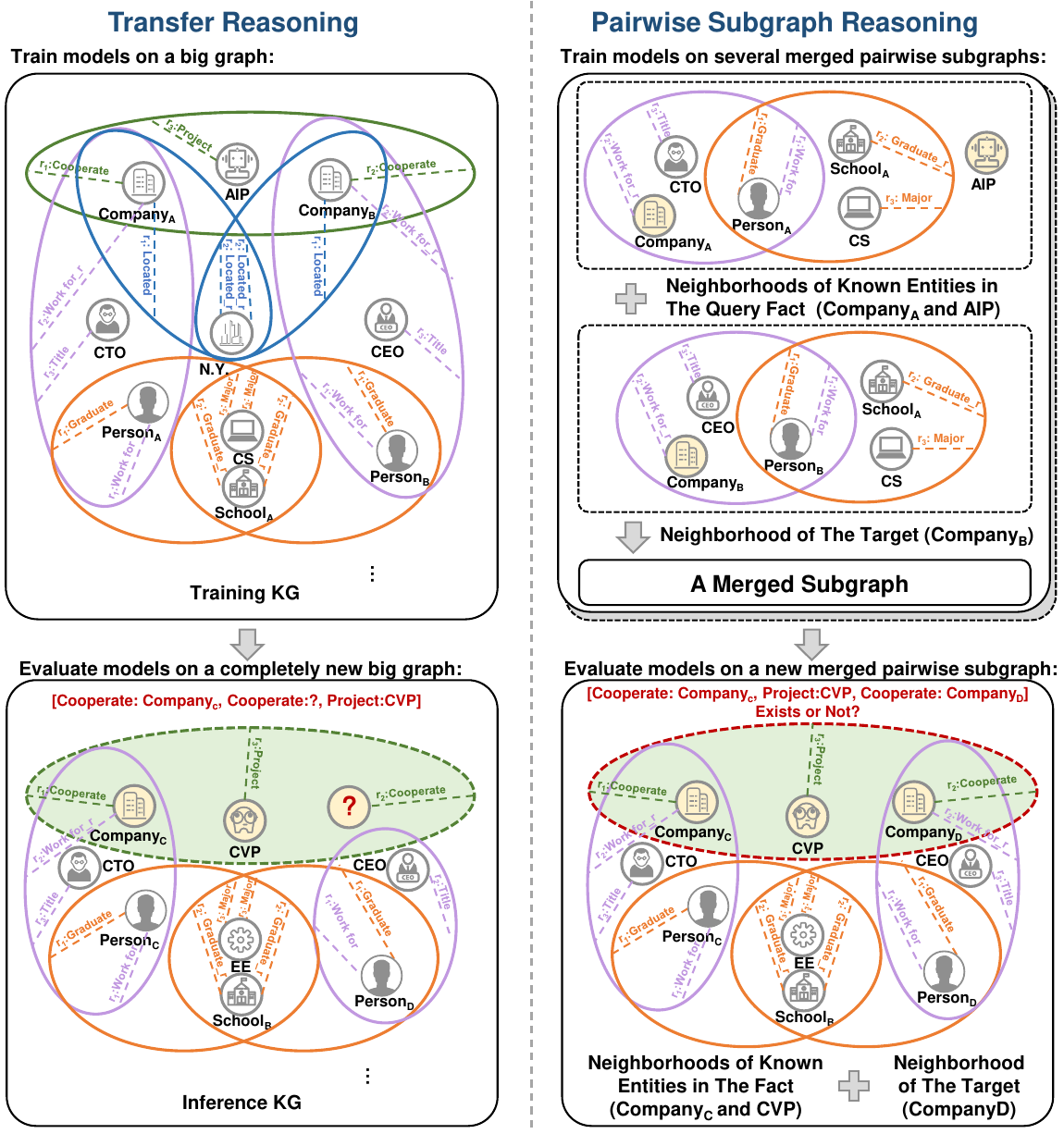}
	\caption{Task Examples. In transfer reasoning tasks, models are trained on one graph and then predict missing links among candidate entities in another inference graph (with no entity overlapping). In pairwise subgraph reasoning tasks, models are given the merged neighborhood of a target entity and known entities of an incomplete fact to assess the likelihood that the target completes the incomplete fact.}
	\label{fig_task}
\end{figure}

\paragraph{\textbf{Inspiration.}}
In fully inductive scenarios, models need to discern entity-independent patterns for making predictions in completely new graphs. This has proven challenging for previous embedding-based methods in binary ILP tasks \citep{ref19}. In n-ary ILP tasks, models face more challenges: \textbf{(i)} The n-ary relational patterns are far more complex, making them difficult to represent using explicit logical rules. \textbf{(ii)} Semantic correlations exist both within individual facts and across different facts, requiring models to effectively capture multi-hop correlations. For instance, as illustrated in Figure \ref{fig_task}, when predicting the cooperation relation, models need to learn that "companies with management acquaintances are more likely to cooperate". This rule involves multiple across-fact n-ary relations, and is hard to represent using explicit logic. In triple-based KGs, GNN-based subgraph reasoning methods \citep{ref25,ref50} have achieved significant success by implicitly capturing logical rules in subgraphs. However, they are bi-edge-based and thus inadequate for handling n-ary relations. A promising idea is to use hypergraphs to represent n-ary relational KGs and employ hypergraph neural networks (HGNNs) \citep{ref20} to capture subgraph information. However, this idea faces two key challenges: \textbf{(i)} Existing hypergraph structures fall short in \textbf{representing the semantic relations (also known as roles)} of entities within a fact \citep{ref22}. \textbf{(ii)} Current HGNNs struggle to \textbf{model complex semantic n-ary correlations} \citep{ref31} inner and across facts. 

\paragraph{\textbf{Solution.}}
To address these limitations and enable effective inductive link prediction (ILP) for n-ary relational facts, we propose a solution with \textbf{these two key contributions:}
\begin{itemize}
\item \textbf{The notion of n-ary semantic hypergraph.}
To overcome the first limitation, we introduce the n-ary semantic hypergraph, \textbf{a novel graph structure} rooted in the key-value pair fact representation. Unlike hyper-relational KGs, which are still bi-edge-based, this structure is a direct generalization of traditional hypergraphs. As shown in Figure \ref{fig_task}, each n-ary relational fact is represented as a hyperedge connecting entities through their \textit{semantic relations/roles}. This enables straightforward exploration of any entity's neighborhood (including the qualifiers) and expressing n-ary relations without information loss.
 
\item \textbf{N-ary subgraph aggregating networks.}
To address the second limitation, we develop an n-ary subgraph aggregating network reasoning over the proposed graph structure. Though transductive methods have succeeded in modeling n-ary semantic interactions via Transformer decoders \citep{ref6,ref21}, they are limited to intra-fact interactions. In contrast, we propose the \textbf{N}-ary \textbf{S}emantic \textbf{H}ypergraph \textbf{A}ggregator based on \textbf{R}elational \textbf{T}ransformers (\textbf{NS-HART}). NS-HART bridges this gap by introducing the two-stage message-passing mechanism of HGNNs and leveraging a Transformer with a role-aware encoding mechanism as the aggregating function.
\end{itemize}

Using this framework, NS-HART can capture and utilize inductive clues in the multi-hop neighborhood to make inferences. Moreover, we provide a theoretical analysis from the perspective of score function optimization, explaining the superior inductive capabilities of NS-HART compared to previous transductive methods and HGNNs with existing aggregating functions.

\paragraph{\textbf{Experiments.}}
Building on pioneering inductive n-ary work \citep{ref16} and triple-based KG works \citep{ref19}, we consider three realistic fully inductive LP scenes and introduce a series of benchmark tasks: transfer reasoning with and without entity features, and pairwise subgraph reasoning. We evaluate the performance of traditional triple-based methods, hyper-relational-based methods, and our n-ary subgraph reasoning-based methods. The results highlight the superiority of the n-ary subgraph reasoning framework and the exceptional inductive ability of NS-HART.

\section{Related Work}
\paragraph{\textbf{Transductive Link Prediction on N-ary Relational KGs.}}
Most existing link prediction methods for n-ary relational KGs are in transductive settings and are embedding-based \citep{ref9,ref10,ref12,ref15}. They focus on learning latent embeddings for entities and relations by optimizing a plausibility score function. When devising score functions for n-ary facts, key-value pair representation-based methods often include role-specific or positional embedding mechanisms\citep{ref10,ref12}, while hyper-relational representation-based methods process the main triple and the qualifier pairs separately\citep{ref9,ref15} before merging them. Recently, some pioneering works \citep{ref6,ref21, ref52} have adopted the Sequence Transformer as a decoder to model complicated interactions among intra-fact elements and yield promising outcomes. Furthermore, more recent studies \citep{ref13,ref14}  have leveraged multi-hop graph neighborhood information to enhance entity embeddings before putting them into the Transformer. However, these methods still compress neighboring information into shallow entity embeddings, making them unsuitable for handling new entities\citep{ref19}. In contrast, our solution is inherently inductive, capturing entity-independent patterns via powerful aggregating networks.
	
\paragraph{\textbf{Inductive Link Prediction on Triple-Based KGs.}}
For inductive link prediction without entity features, methods can be typically classified into rule-based or inductive embedding-based. Rule-based methods derive probabilistic logical rules by pattern mining \citep{ref17,ref18}, which can be complex for n-ary facts. Inductive embedding-based methods generally use GNNs \citep{ref25, ref55} to implicitly deduce inferencing rules. Moreover, in inductive scenarios where entity features (textual information) are available, models are tasked with learning from both the topological structure and the node attributes. Given that GNN-based methods tend to seek a comprise between the two, resulting in suboptimal performance \citep{ref27}, many studies have shifted to Transformer architectures \citep{ref28,ref30}. These architectures are capable of encoding relational correlations and textual descriptions simultaneously. Motivated by these, our solution introduces the Transformer architecture in the message-passing process, uniformly handling inductive scenes with and without features.
	
\paragraph{\textbf{GNNs-based Hypergraph Learning.}}
Hypergraphs are widely used to model high-order relations involving more than two entities, with each hyperedge representing a set of nodes. Inspired by the success of GNNs, several works have attempted to extend GNNs to hypergraphs. The prevalent methods are based on hypergraph star expansion \citep{ref22,ref31}, where hyperedges are viewed as virtual nodes, resulting in a bipartite heterogeneous graph. This allows for a two-stage spatial-based HGNN process: the first stage aggregates information from nodes to hyperedges, and the second aggregates from hyperedges to nodes. However, these methods struggle to capture intra-edge semantic relations within hypergraphs. Some works refer to "relational hypergraph" \citep{ref54} and “heterogeneous hypergraph” \citep{ref33,ref34} to denote hyperedges with semantics. In these hypergraphs, semantics are assigned solely to the hyperedge (also known as main relations or types), overlooking the semantic roles of the entities within. Diverging from these, we propose a more expressive structure where semantic roles exist between entities and hyperedges, specifically designed for n-ary relational facts.

\section{Task Description}
In triple-based KGs, there are generally two types of fully ILP tasks, i.e. transfer reasoning and pairwise subgraph reasoning \citep{ref56}. In this work, we extend these tasks to n-ary KGs to handle different realistic scenes. As shown in Figure \ref{fig_task}, \textbf{the given context information, goal, and inductive ability focus} differ for each sub-task:

\paragraph{\textbf{(i) Transfer Reasoning with Entity Features (TR-EF)}}
Following the setup in \citep{ref16}, this task involves training models on one graph and evaluating them on a distinct inference graph. The inference graph has a non-overlapping set of entities but retains the same set of relations as the training graph. The goal is to predict potential links among all entities in the inference graph, specifically predicting the missing entity "$\textbf{?}$" for an incomplete fact $ \left[ r_1:v_1,...,r_p:\textbf{$?$},...,r_n:v_n \right]$. \textbf{Note that, entity features are provided in this task,} focusing on models' inductive capabilities in exploiting both structural and feature information.
\paragraph{\textbf{(ii) Transfer Reasoning with No Entity Features (TR-NEF)}}
This sub-task is \textbf{under the same setting as TR-EF but without entity features}. It emphasizes models' intrinsic inductive reasoning ability based on structural patterns, which is crucial in situations with limited attribute information.
\paragraph{\textbf{(iii) Pairwise Subgraph Reasoning (PSR)}}
For an incomplete fact $ \left[ r_1:v_1,...,r_p:\textbf{$?$},...,r_n:v_n \right]$ and a target entity $v_t$, we are given a subgraph containing the neighborhoods of known entities within this fact and the neighborhood of the target entity. The goal is to predict the likelihood that the target entity fits the missing place given the neighborhood information. This sub-task focuses on logically inferring link probability solely based on the local subgraph structure, as discussed in~\cite{ref19,ref25}. It's particularly useful for predicting the likelihood of specific links in large graph scenarios.

\begin{figure*}\centering
	\includegraphics[width=0.8\linewidth]{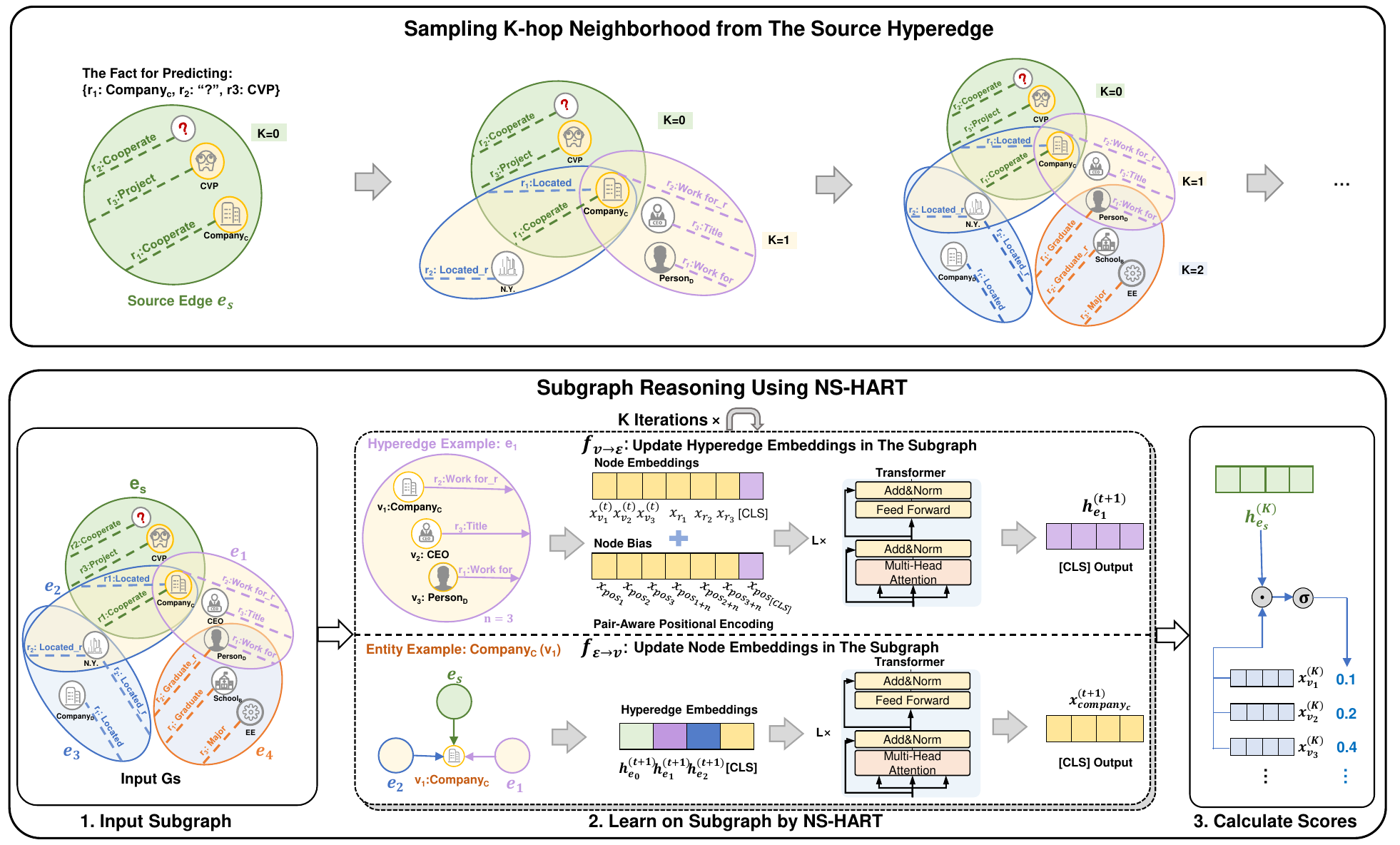}
	\caption{The overall structure of the proposed learning framework. To illustrate the message passing processes of NS-HART, we take updating embeddings of hyperedge $e_1$ and entity $v_1$ at iteration t as an example.}
	\label{fig_main_framework}
\end{figure*}

\section{Methodology}
Based on the assumption that extended neighboring subgraphs of the \textbf{source entities (i.e. known entities in the incomplete fact being predicted)} contain inference clues for ILP \citep{ref25}, we aim to develop an effective solution tailored for n-ary relational KGs via subgraph reasoning. The main framework is shown in Figure \ref{fig_main_framework}. First, we propose the notion of n-ary semantic hypergraph and describe the neighborhood sampling process \textbf{(Sec 4.1)}. Then, we introduce n-ary subgraph aggregating networks to learn from the sampled subgraph, deriving node and hyperedge embeddings enriched with neighborhood contexts \textbf{(Sec 4.2)}. Finally, link prediction is achieved by calculating scores between the updated embeddings of the source hyperedge and the potential candidate entities \textbf{(Sec 4.3)}.

\subsection{The N-ary Semantic Hypergraph}
\paragraph{\textbf{Defining N-ary Semantic Hypergraphs.}}
Let $\mathcal{G}=\left( \mathcal{V},\mathcal{E} \right)$ denote a traditional hypergraph without any semantic information, where $\mathcal{V}$, $\mathcal{E}$  are the node set and the hyperedge set, respectively. Each hyperedge is a set of multiple nodes (at least two or more). For a traditional hyperedge $\widetilde{e} \in \mathcal{E} $ containing $n$ nodes, it can be denoted as $ \widetilde{e} = \left\{ v_i | v_i\in \mathcal{V} \right\} _{i=1}^{n} $.

The proposed semantic hypergraph $\mathcal{G}^*=\left( \mathcal{V},\mathcal{E},\mathcal{R} \right)$ is defined as a generalized hypergraph by \textbf{pairing entities with their semantic relations within each hyperedge}, where $\mathcal{R}$ is the semantic relation set. This can be easily implemented by altering the original boolean value in the hypergraph adjacency matrix to represent specific semantic relations. For a semantic hyperedge  $ e\in\mathcal{E} $ containing $n$ nodes, it can be denoted as $ e=\left\{ \left( r_i, v_i \right) |r_i\in \mathcal{R},v_i\in \mathcal{V} \right\} _{i=1}^{n} $. Here, each $v_i$ represents an entity, and $r_i$ denotes its \textbf{roles (also known as relations)} within the n-ary fact. Each element in the adjacency matrix ${H}\in\mathcal{R}^{|\mathcal{V}|\times |\mathcal{E}|}$ of the hypergraph can be denoted as:
\begin{equation}
H\left( v,e \right) =\left\{ \begin{array}{l} 	r_{id},\ if\ \left( r, v \right) \in e\\ 	0,\ otherwise\\ \end{array} \right.
\end{equation}
Where $r_{id}$ denotes the mapping value of the semantic role $r$. Note that, in very few cases, an entity may have multiple roles in a fact. To accommodate such situations, we can also extend the element in $H$ to a mapping value of a set of relations.

In practice, the proposed n-ary semantic hypergraph can be seamlessly transformed into a bipartite heterogeneous graph with two types of nodes: entity nodes and virtual hyperedge nodes, using its star expansion form \citep{ref59}. Compared to alternative graph representations for n-ary facts (e.g. hyper-relational KGs \citep{ref6}, relational hypergraphs \citep{ref54}), the proposed structure offers several distinct advantages (especially for fully ILP), such as \textbf{more expression flexibility, facilitating neighborhood expansion, and synergy with HGNNs.} (See Appendix \ref{sec:NS-Graph} for more details).
	
\paragraph{\textbf{Defining Neighborhoods for Nodes.}}
To better describe the neighborhood sampling process, we first provide the specific definitions related to a node’s neighborhood below. In this context, \textbf{“neighborhood”} refers to a set of hyperedges, and \textbf{"neighbor nodes"} refers to a set of nodes.

\paragraph{Definition (Intra-Edge Neighbor Nodes)}
We define the set of nodes within a hyperedge $ e\in \mathcal{E} $ as follows: $\mathscr{N}(e) = \left\{ v|\left( r,v \right) \in e \right\}$. For a node $v_i$ , its intra-edge neighbor nodes for any of its belonging hyperedge $e$ are represented as: $\mathscr{N} ( v_i,e ) = \mathscr{N}(e) \setminus \{v_i\}$.

\paragraph{Definition (K-hop Neighborhood)}
First, we define the 1-hop neighborhood of $v_i$ as the set of hyperedges including $v_i$, denoted as: $\mathcal{H}_1\left( v_i \right) =\left\{ e|\left( r_{\left( v_i,e \right)},v_i \right) \in e \right\}$. To further obtain the K-hop neighborhood of a node, we need to get nodes in its (K-1)-hop neighborhood first, and then union these nodes' 1-hop neighborhoods. The K-hop neighborhood ($k\ge2$) of a node $v_i$ is derived as follows: $\mathcal{H}_k\left( v_i \right)=\bigcup_{u\in\mathscr{N}\left(e\right), e\in \mathcal{H}_{k-1}\left(v_i\right)}{\{\mathcal{H}_1\left(u\right)}\}$.
	
\paragraph{\textbf{Sampling Input Subgraphs.}}
As shown in Figure~\ref{fig_main_framework}, given the source hyperedge $e_s$ representing an incomplete fact, we treat the predicting target \textbf{"$?$"} as a virtual entity and construct the input subgraph $G_S$ by merging K-hop neighborhoods of nodes in $\mathscr{N}\left( \textbf{?},e_s \right)$. For PSR tasks, we also extract another subgraph containing K-hop neighborhoods of the target entity $v$ and merge the two subgraphs as the input subgraph. Given that the entire neighborhood can expand exponentially, we implement a simple sampling strategy. When constructing the 1-hop neighborhood, we sample $m$ adjacency hyperedges from each node; for the next K-hop neighborhood, we sample $\log_k(m)$ adjacency hyperedges from each node in ($k-1$)-hop neighborhood. This strategy efficiently saves computational resources and mitigates overfitting risk. In practice, we store the hypergraph as HeteroData in PyTorch Geometric \footnote{\url{https://pytorch-geometric.readthedocs.io}} and employ NeighborLoader \citep{ref41} for efficient and cached batch-wise sampling.

\subsection{Subgraph Reasoning Using NS-HART}
Given the n-ary subgraph, our aim is to develop a neural network that aggregates information and captures patterns in a data-driven manner. Two main issues need to be considered: \textbf{(i) Preserving multi-hop hypergraph structural information. (ii) Modeling interactions between relations and entities.}

For the first issue, previous spatial-based HGNNs \citep{ref22,ref31,ref32} have proposed a good solution. They often employ a two-stage message-passing procedure to aggregate information from vertex to hyperedge, and hyperedge to node, respectively. The propagation rules can be unified within a composition of two multiset functions \citep{ref31}. Let $\mathcal{F}$ denote a multiset function that is permutation invariant, the message-passing process can be defined as:
\begin{equation}
	\left\{ \begin{array}{l} 	h_{e}^{\left( t+1 \right)}=\mathcal{F}_{\mathcal{V}\rightarrow \mathcal{E}}\left( \left\{ x_{j}^{\left( t \right)} \right\} _{j\in e};h_{e}^{\left( t \right)} \right)\\ 	x_{v}^{\left( t+1 \right)}=\mathcal{F}_{\mathcal{E}\rightarrow \mathcal{V}}\left( \left\{ h_{i}^{\left( t+1 \right)} \right\} _{v\in i};x_{v}^{\left( t \right)} \right)\\ \end{array} \right.
\end{equation}
where $\mathcal{F}_{\mathcal{V}\rightarrow \mathcal{E}}$ and $\mathcal{F}_{\mathcal{E}\rightarrow \mathcal{V}}$ are two multiset functions, $x^{\left( t\right)}$ and $h^{\left( t\right)}$ denote the hidden embedding of a node and a hyperedge at iteration $t$, respectively. Thus, selecting appropriate multiset functions based on specific tasks becomes crucial. For instance, UniSAGE and UniGAT  \citep{ref22} generalize aggregation functions in traditional GNNs to hypergraphs, while AllSetTransformer \cite{ref31} employs set Transformer \cite{ref35} as multiset functions to learn within each set.

Though previous HGNNs succeed in learning on traditional hypergraphs, they struggle in modeling interactions between roles and entities. Inspired by the \textbf{Transformer's ability to model complex interactions within sets \citep{ref35,ref45} and semantic correlations in arbitrarily long sequences}, we use a Sequence Transformer as the backbone for multiset functions and introduce a new graph aggregating network, NS-HART. Next, we will detail its message-passing process, including the $\mathcal{V}\rightarrow \mathcal{E}$ and the $\mathcal{E}\rightarrow \mathcal{V}$ steps.

\paragraph{\textbf{$\mathcal{V}\rightarrow \mathcal{E}$ Process.}}
This process targets at modeling complicated interactions within a hyperedge, resulting in updated hyperedge embeddings implicitly endowed with intra-edge information. When choosing the multiset functions, two types of interactions need to be taken into account, i.e. the mutual interactions among entities within a hyperedge, and pairwise connections between each entity and its associated role. To address this challenge, we propose a simple yet effective solution: view both entities and semantic roles as tokens, mark pairwise role-entity connections as node bias, and then employ the Transformer as the multiset function for aggregation. Specifically, given a hyperedge $ e=\left\{ \left( r_1,v_1 \right) ,...,\left( r_n,v_n \right) \right\} $, the $\mathcal{V}\rightarrow \mathcal{E}$ aggregation operates as follows.

First, we develop a \textbf{role-aware positional encoding mechanism} to indicate the mapping relationship between roles and their corresponding entities. For a hyperedge containing $n$ pairs, we initially assign a positional integer to each entity. Here we randomly assign 1 to $n$ (with no repeating) to mark each entity’s position as $ pos_{v_i} $. For each semantic role $ r_i $, we assign its positional mark based on its associated entity, denoted as $pos_{r_i}=pos_{v_i}+n$. For the source hyperedge, we highlight the missing entity’s role using a unique positional indicator. As both entities and semantic roles are viewed as tokens, we update the original token embeddings by adding its positional encodings: $x_{\tilde{i}}^{\left( t \right)}\gets x_{\tilde{i}}^{\left( t \right)}+x_{pos_{\tilde{i}}}$, where $ \tilde{i} $ denotes an entity or a relation in the hyperedge, $x_{\tilde{i}}$ and $x_{pos_{\tilde{i}}}$ denotes its $d$-dimensional learnable hidden embedding and positional embedding, respectively. Note that, this mechanism can also be main-qualifier aware by setting positional marks to distinguish entities as either primary or qualifiers.

Next, we add a "[CLS]" token with a distinct positional integer to the token sequence, which is a common practice in Transformer-based text classification to readout the sentence information \citep{ref37}. Finally, we feed the sequence embedding into a Transformer and take the output corresponding to the "[CLS]" token as the updated hyperedge embedding. We can express the {$\mathcal{V}\rightarrow \mathcal{E}$ process} as: 
\begin{equation}
	h_{e}^{\left( t+1 \right)}=\,\,Trans_{\left[ CLS \right]}\left( x_{\left[ CLS \right]},x_{r_1},x_{v_1}^{\left( t \right)},... \right)
\end{equation}
	
\paragraph{\textbf{$\mathcal{E}\rightarrow \mathcal{V}$ Process.}}
This process aims to update the entity embeddings in the subgraph by aggregating information from its belonging hyperedges. To ensure the expressive capability and the equivariant feature of the aggregating networks, we also choose Transformer as the backbone for the multiset function. During this process, we don’t add the entities' semantic roles to the input sequence again, as they have already been contained in updated hyperedge embeddings. Given an entity and its 1-hop neighborhood, the $\mathcal{E}\rightarrow \mathcal{V}$ process is denoted as: 
\begin{equation}
	x_{v}^{\left( t+1 \right)}=\,\,Trans_{\left[ CLS \right]}\left( \left[ x_{\left[ CLS \right]},h_{e_1}^{\left( t+1 \right)},... \right] \right)
\end{equation}

Through K iterations of $\mathcal{V}\rightarrow \mathcal{E}$ aggregations and $\mathcal{E}\rightarrow \mathcal{V}$ aggregations, the subgraph aggregating network can broaden attention to the K-hop neighborhoods. 

\subsection{Score Calculating}
For each subgraph, we use subgraph aggregating networks to update the embeddings of the entities and hyperedges in it. The link prediction scores $\mathcal{P}$ can be calculated by \textbf{inner products between the updated embeddings of source hyperedges and the candidate entities}, expressed as:
\begin{equation}
	\mathcal{P}\left( e_s,\hat{V}_{} \right) =\sigma \left( h_{e_s}^{(K)} ·X_{\hat{V}}^{\left( K \right)} \right)
\end{equation}

Where $\hat{V}$ is the set of candidate entities, and $X_{\hat{V}}^{\left( K \right)}$ is the matrix consisting of their embeddings (each row as $x_{v_i}^{( K )}$) after K iterations. Note that, for entities not in the subgraph in transfer reasoning tasks, we retain their embeddings as the initial ones. The overall training process can be found in Appendix~\ref{sec:training}.

\section{Theoretical Analysis}
In this section, We first analyze the subgraph reasoning framework from the score function optimization perspective. Based on this perspective, we compare NS-HART with other methods and shed light on its superiority for n-ary ILP tasks.

\subsection{Optimization Perspective of The Subgraph Reasoning Framework}
As summarized in previous studies \cite{ref56}, most neural-based LP methods for KGs actually optimize a plausibility score function. For n-ary relational KGs, the score function for each query fact $\mathcal{Q} = [\{\left(r_i: v_i\right)\}_{i=1}^n]$ (with $v_p$ as the missing entity) can be denoted as $\mathscr{S}(\mathcal{Q}, v_p, G_S)$, where $G_S$ represents the neighborhoods. Next, we will deduce the specific formation of score functions when using a subgraph reasoning framework.

As depicted in \textbf{Sec 4.3}, when using the subgraph reasoning framework, the score function can be roughly denoted as:
\begin{equation}
\mathscr{S} = h_{e_s}^{(K)}·x_{v_p}^{(K)}
\end{equation}

In $\mathcal{V}\rightarrow \mathcal{E}$ process, all hyperedge embeddings in the subgraph are updated as:
\begin{equation}
h_e^{ (t+1)} = \mathcal{F}_{\mathcal{V}\rightarrow \mathcal{E}}(\{x_{r_1},x_{v_1}^{( t )},...,x_{r_n},x_{v_n}^{( t )}\})
\end{equation}

In $\mathcal{E}\rightarrow \mathcal{V}$ process, all entity embeddings in the subgraph are updated as:
\begin{equation}
x_{v}^{(t+1)} =\mathcal{F}_{\mathcal{E}\rightarrow \mathcal{V}}( \{ h_{1}^{( t+1 )},...,h_{m}^{( t+1 )} \} )
\end{equation}

In real applications, we set $\mathcal{F}_{\mathcal{E}\rightarrow \mathcal{V}}$ and $\mathcal{F}_{\mathcal{V}\rightarrow \mathcal{E}}$ to the same multiset function $\mathcal{F}$. By iteratively substituting Equation (7) and (8) to represent $h_{e_s}^{(K)}$ and $x_{v_p}^{(K)}$, we can get (take $h_{e_s}^{(K)}$ as an example):
\begin{align}
	h_{e_s}^{(K)} &=\mathcal{F}(\{ x_{r_1},x_{v_1}^{( K-1)},...,x_{r_n},x_{v_n}^{( K-1 )} \}) \\
	& =\mathcal{F}(\{x_{r_1}, \mathcal{F}(h_{1_1}^{( K-1)},...,h_{1_m}^{( K-1)}),...\} )\\     & = ... \\
        & = \mathscr{F}\left( \left\{ x_{v_i}^{(0)}, x_{r_i} \right\}_{i=1}^M \mid (v_i, r_i) \in e_j, e_j \in G_S \right)
\end{align}
where $\mathscr{F}$ \textbf{is a nested function} and $M$ is the number of pairs in the subgraph. In this way, $h_{e_s}^{(K)}$ and $x_{v_p}^{(K)}$ can be represented using the initial embeddings of entities and relations in $G_S$. Thus, the score function for the subgraph reasoning framework is deduced as:
\begin{equation}
\mathscr{S} = \mathscr{F}_{\mathcal{Q}}\left(\left\{ x_{v_i}^{\left( 0 \right)}, x_{r_i} \right\}_{i=1}^M \right) \cdot \mathscr{F}_{v_p}\left(\left\{ x_{v_i}^{\left( 0 \right)}, x_{r_i} \right\}_{i=1}^M \right)
\end{equation}

Here, $\mathscr{F}_{*}(\cdot)$ denotes the function $\mathscr{F}$ with $*$ representing the output and $\cdot$ representing the input. Each $( v_j,r_j)$ denotes a pair in $G_S$. From Equation (13), it is evident that \textbf{we use $\mathscr{F}$ with nested aggregating functions $\mathcal{F}$ to model complicated interactions in the subgraph}. This approach stores useful knowledge in the aggregating network itself, rather than in the limited-dimensional entity embeddings. This design enhances inductive performance and aligns with principles from previous inductive research, such as NodePiece \citep{ref58}. Therefore, the expressive power of $\mathcal{F}$ is a key factor influencing the model's inductive ability.

\begin{table}
  \centering
  \caption{Score functions of NS-HART and other methods.}
  \resizebox{0.5\textwidth}{!}{
    \begin{tabular}{cccc}
    \toprule
    & \textbf{Score Function} & \textbf{$\mathscr{F}$} & \textbf{$\mathcal{F}$}\\
    \midrule
    \multirow{2}[1]{*}{NS-HART} & \multirow{2}[1]{*}{$\mathscr{F}_{\mathcal{Q}}(G_s) \cdot \mathscr{F}_{v_p}(G_s)$}  & \multirow{2}[1]{*}{$\mathcal{F}(\mathcal{F}(...),...)$} & Relational Sequence\\
    & & & Transformer\\
    \midrule
    \multirow{2}[1]{*}{Other} & \multirow{3}[1]{*}{$\mathscr{F}_{\mathcal{Q}}(G_s) \cdot \mathscr{F}_{v_p}(G_s)$}  & \multirow{3}[1]{*}{$\mathcal{F}(\mathcal{F}(...),...)$} & MLPs\\
    & & & Attentions\\
    HGNNs& & & Set Transformer\\
    \midrule
    Only & \multirow{2}[1]{*}{$\mathscr{F}_{\mathcal{Q}}(\mathcal{Q}) \cdot x_{v_p}$} & \multirow{2}[1]{*}{Transformer} & \multirow{2}[1]{*}{-}\\
    Decoder  &   &  & \\
    \midrule
    GNN+ & \multirow{2}[1]{*}{$\mathscr{F}_{\mathcal{Q}}(\mathcal{F}_{\mathcal{Q}}(G_S)) \cdot x_{v_p}$}  & \multirow{2}[1]{*}{Transformer} & \multirow{2}[1]{*}{GNN}\\
    Decoder  &   &  & \\
    \bottomrule
    \end{tabular}%
    }
  \label{tab_scores}%
\end{table}%

\subsection{NS-HART vs. Other Methods}
We summarize score functions of n-ary LP methods in Table \ref{tab_scores}. Among them, NS-HART and HGNNs are based on the subgraph reasoning framework, "Only Decoder" \citep{ref16} and "GNN+Decoder" \citep{ref6,ref13,ref14} methods are popular in transductive settings. Based on these analyses, we provide the following comparisons:

\paragraph{\textbf{NS-HART vs. Transductive Embedding-Based Methods}} "Only decoder" methods \citep{ref16} often use Transformer decoders to model intra-edge interactions. When the $\mathcal{V}\rightarrow \mathcal{E}$ process is performed only once, NS-HART approximately degrades to these methods, with $\mathscr{F}$ degrading to a no nested Transformer. Through multi-hop aggregating, NS-HART directly captures larger-scale information. "GNN+Decoder" methods first employ GNN to enrich node embeddings before applying Transformer decoders \citep{ref13,ref14}. In terms of the score function, $\mathscr{F}$ remains a single Transformer, with $GNN_{\mathcal{Q}}(G_s)$ as the input. Since GNNs compress multi-hop information into low-dimensional embeddings, their expressive power is limited. In contrast, NS-HART integrates Transformers into the message-passing process, resulting in a powerful end-to-end subgraph aggregator.

\paragraph{\textbf{NS-HART vs. Other HGNNs}} As discussed above, the expressive power of $\mathcal{F}$ is crucial for a model's inductive ability. In NS-HART, $\mathcal{F}$ is a role-aware Transformer. Other HGNNs use different mechanisms: UniSAGE uses the sum aggregator, UniGAT uses the attention mechanism \citep{ref22}, and AllSetTransformer \citep{ref31} uses the Set Transformer \citep{ref35}. Compared to NS-HART, they struggle to handle pairwise role-entity correlations and generally have less expressive power than the Sequence Transformer.

In Appendix \ref{sec:NS-HART}, we provide more analysis of NS-HART, including its efficiency and its comparison to graph Transformers.

\section{Experiments}
Focusing on fully ILP for n-ary relational facts, we conduct extensive experiments to assess various models and demonstrate the superiority
of learning on n-ary semantic hypergraphs with NS-HART. Specifically, we delve into the following questions:
\textbf{(Q1) Performance Comparision.} How does NS-HART perform against baseline models?
\textbf{(Q2) Ablation Study.} What is the impact of K-hop message-passing and high-order correlations modeling?
How effective is the role-aware positional encoding mechanism in NS-HART?
\textbf{(Q3) Key Parameter Analysis.} How do parameters like sampling scale and Transformer layers affect performance?
\textbf{(Q4) Case Study.} Take an example to visualize attention scores for entities and hyperedges in the subgraph.

\subsection{Experimental Setup}
\paragraph{\textbf{Datasets.}}
To measure the benefits of employing hyper-relational representations on ILP tasks, \citep{ref16} introduced a dataset named WD20K. To investigate ILP tasks with entity features, we employ the fully inductive version of WD20K for TR-EF in our experiments, namely WD20K (100) and WD20K (66), where \textbf{the values in parentheses indicate the proportions of n-ary relational facts}. Moreover, each of these datasets has two versions, where V1 has a larger training graph and V2 has a bigger inference graph. For TR-NEF tasks, we introduce additional datasets, FI-MFB (100) and FI-MFB (33), derived from the n-ary knowledge base M-FB15K \citep{ref10}. For PSR tasks, we sample the 2-hop neighborhood around nodes within the source hyperedge and the target node to form each input graph. Details of the datasets can be found in Appendix \ref{sec:dataset}.

\paragraph{\textbf{Compared Methods.}}
We evaluate the following cates of methods: \textbf{triple-based methods (BLP, CompGCN), hyper-relational-based methods (QBLP, StarE, GRAN), HGNNs using other multiset functions (UniSAGE, UniGAT, AllSetTransformer), and NS-HART}. In addition, we add NS-HART (intra-edge) as an ablation study variant of NS-HART, which performs $\mathcal{V}\rightarrow \mathcal{E}$ process only once. For PSR tasks, our evaluation is limited to HGNNs and NS-HART, given the necessity for models to directly operate and logically infer on subgraphs. We explain the reasons for choosing these baselines and their details in Appendix \ref{sec:baseline}.

\paragraph{\textbf{Evaluation Settings}}
For TR-EF and TR-NEF tasks, we evaluate model performance using two widely adopted ranking metrics, the mean reciprocal rank (MRR) and the hits rate HITS@10. Specifically, for each n-ary fact with one missing entity, we rank the true target among entities from the filtered candidate set. For PSR tasks, we choose a random negative entity from the input subgraph for each fact and report the AUC-PR results. Since the entities in the facts of existing KGs mainly distinguish between primary (i.e. the head and the tail entity) and qualifier parts, we report the results of predicting primary entities and qualifier entities respectively in most experiments. In our experiments, all models are trained and tested for 3 times with different random seeds, and the mean performance is reported.

\paragraph{\textbf{Other Settings.}}
Full details of parameter settings are included in Appendix \ref{sec:param}.

\begin{table*}
	\large
	\centering
	\caption{Results of TR-EF tasks. Methods with * of primary entity prediction results are reported using the best results from  \citep{ref16}.} 
	\resizebox{\textwidth}{!}{
		\begin{tabular}{c|cc|cc|cc|cc|cc|cc|cc|cc}
			\toprule
			\multirow{3}[1]{*}{\textbf{Method}} & \multicolumn{4}{c|}{\textbf{WD20K (100) V1}} & \multicolumn{4}{c|}{\textbf{WD20K (66) V1}} & \multicolumn{4}{c|}{\textbf{WD20K (100) V2}} & \multicolumn{4}{c}{\textbf{WD20K (66) V2}} \\
			\cmidrule(r){2-17}
			& \multicolumn{2}{c|}{\textbf{Primary}} & \multicolumn{2}{c|}{\textbf{Qualifier}} & \multicolumn{2}{c|}{\textbf{Primary}} & \multicolumn{2}{c|}{\textbf{Qualifier}} & \multicolumn{2}{c|}{\textbf{Primary}} & \multicolumn{2}{c|}{\textbf{Qualifier}} & \multicolumn{2}{c|}{\textbf{Primary}} & \multicolumn{2}{c}{\textbf{Qualifier}} \\
			\cmidrule(r){2-3} \cmidrule(l){4-5} \cmidrule(l){6-7} \cmidrule(l){8-9} \cmidrule(l){10-11} \cmidrule(l){12-13} \cmidrule(l){14-15} \cmidrule(l){16-17}
			& \textbf{MRR} & \textbf{HITS@10} & \textbf{MRR} & \textbf{HITS@10} & \textbf{MRR} & \textbf{HITS@10} & \textbf{MRR} & \textbf{HITS@10} & \textbf{MRR} & \textbf{HITS@10} & \textbf{MRR} & \textbf{HITS@10} & \textbf{MRR} & \textbf{HITS@10} & \textbf{MRR} & \textbf{HITS@10} \\
			\midrule
			\textbf{BLP*} & 0.057 & 0.123 & 0.078 & 0.115 & 0.021 & 0.044 & 0.187 & 0.222 & 0.040  & 0.092 & 0.158 & 0.200 & 0.016 & 0.034 & 0.062 & 0.085 \\
			\textbf{CompGCN*} & 0.104 & 0.184 & 0.144 & 0.251 & 0.058 & 0.128 & 0.134 & 0.199 & 0.026 & 0.053 & 0.118 & 0.202 & 0.026 & 0.045 & 0.065 & 0.108 \\
			\midrule
			\midrule
			\textbf{QBLP*} & 0.107 & 0.245 & 0.188 & 0.305 & 0.043 & 0.093 & 0.281 & 0.377 & 0.067 & 0.12  & 0.26  & 0.347 & 0.021 & 0.049 & 0.148 & 0.211 \\
			\textbf{StarE*} & 0.113  & 0.213  & 0.192  & 0.270  & 0.068  & 0.134  & 0.212  & 0.277  & 0.051  & 0.129  & 0.236  & 0.311  & 0.051  & 0.098  & 0.083  & 0.125  \\
			\textbf{GRAN} & 0.098  & 0.202  & 0.153  & 0.242  & 0.034  & 0.076  & 0.257  & 0.328  & 0.053  & 0.108  & 0.255  & 0.377  & 0.019  & 0.037  & 0.121  & 0.195  \\
			\midrule
			\midrule
			\textbf{UniSAGE} & 0.141 & 0.360  & 0.242 & 0.485 & 0.089 & 0.167 & 0.189 & 0.396 & 0.115 & 0.188 & 0.197 & 0.281 & 0.028 & 0.053 & 0.064 & 0.121 \\
			\textbf{UniGAT} & 0.147 & 0.303 & 0.125 & 0.284 & 0.065 & 0.176 & 0.178 & 0.319 & 0.077 & 0.158 & 0.087 & 0.168 & 0.047 & 0.100   & 0.067 & 0.166 \\
			\textbf{AllSetTransformer} & 0.121 & 0.264  & 0.098 & 0.220 & 0.056 & 0.121 & 0.169  & 0.238 & 0.051 & 0.143 & 0.074 & 0.164 & 0.033 & 0.094 & 0.076 & 0.122 \\
			\cmidrule(l){1-17}
			\textbf{NS-HART (intra-edge)} & 0.112 & 0.241 & 0.188 & 0.291 & 0.042 & 0.095 & 0.251 & 0.340  & 0.055 & 0.131 & 0.312 & 0.414 & 0.027 & 0.053 & 0.175 & 0.241 \\
			\multirow{2}[1]{*}{\textbf{NS-HART}} & \textbf{0.498} & \textbf{0.646} & \textbf{0.527} & \textbf{0.688} & \textbf{0.196} & \textbf{0.326} & \textbf{0.461} & \textbf{0.711} & \textbf{0.325} & \textbf{0.471} & \textbf{0.401} & \textbf{0.596} & \textbf{0.176} & \textbf{0.267}& \textbf{0.408} & \textbf{0.562} \\
			 & \textbf{±0.06} & \textbf{±0.05} & \textbf{±0.03} & \textbf{±0.03} & \textbf{±0.05} & \textbf{±0.02} & \textbf{±0.03} & \textbf{±0.03} & \textbf{±0.05} & \textbf{±0.06} & \textbf{±0.04} & \textbf{±0.03} & \textbf{±0.02} & \textbf{±0.02}& \textbf{±0.03} & \textbf{±0.04} \\
			\bottomrule
		\end{tabular}%
		\label{tab_main_featurel}%
	}
\end{table*}%

\begin{table*}
	\centering
	\large
	\caption{Results of TR-NEF tasks.}
	\resizebox{\textwidth}{!}{
		\begin{tabular}{c|cc|cc|cc|cc|cc|cc|cc|cc}
			\toprule
			\multirow{3}[1]{*}{\textbf{Method}} & \multicolumn{4}{c|}{\textbf{WD20K (100) V1}} & \multicolumn{4}{c|}{\textbf{WD20K (66) V1}} & \multicolumn{4}{c|}{\textbf{FI-MFB (100)}} & \multicolumn{4}{c}{\textbf{FI-MFB (33)}} \\
			\cmidrule(r){2-17}
			& \multicolumn{2}{c|}{\textbf{Primary}} & \multicolumn{2}{c|}{\textbf{Qualifier}} & \multicolumn{2}{c|}{\textbf{Primary}} & \multicolumn{2}{c|}{\textbf{Qualifier}} & \multicolumn{2}{c|}{\textbf{Primary}} & \multicolumn{2}{c|}{\textbf{Qualifier}} & \multicolumn{2}{c|}{\textbf{Primary}} & \multicolumn{2}{c}{\textbf{Qualifier}} \\
			\cmidrule(r){2-3} \cmidrule(l){4-5} \cmidrule(l){6-7} \cmidrule(l){8-9} \cmidrule(l){10-11} \cmidrule(l){12-13} \cmidrule(l){14-15} \cmidrule(l){16-17}
			& \textbf{MRR} & \textbf{HITS@10} & \textbf{MRR} & \textbf{HITS@10} & \textbf{MRR} & \textbf{HITS@10} & \textbf{MRR} & \textbf{HITS@10} & \textbf{MRR} & \textbf{HITS@10} & \textbf{MRR} & \textbf{HITS@10} & \textbf{MRR} & \textbf{HITS@10} & \textbf{MRR} & \textbf{HITS@10} \\
			\midrule
			\textbf{QBLP} & 0.067 & 0.132 & 0.054 & 0.103 & 0.052 & 0.122 & 0.121 & 0.186 & 0.116 & 0.231 & 0.050  & 0.089 & 0.182 & 0.382 & 0.162 & 0.252 \\
			\textbf{StarE} & 0.079 & 0.189 & 0.053 & 0.105 & 0.044 & 0.097 & 0.117 & 0.183 & 0.132 & 0.311 & 0.054 & 0.094 & 0.219 & 0.403 & 0.111 & 0.192 \\
			\textbf{GRAN} & 0.075 & 0.143 & 0.056 & 0.089 & 0.039 & 0.095 & 0.109 & 0.162 & 0.116  & 0.262 & 0.054 & 0.102 & 0.161 & 0.348 & 0.096 & 0.189 \\
			\midrule
			\midrule
			\textbf{UniSAGE} & 0.125 & 0.316 & 0.156 & 0.349 & 0.058 & 0.138 & 0.148 & 0.281 & 0.039 & 0.068 & 0.031 & 0.082 & 0.127 & 0.240  & 0.088 & 0.186 \\
			\textbf{UniGAT} & 0.202 & 0.366 & 0.178 & 0.347 & 0.092 & 0.181 & 0.109  & 0.303 & 0.099 & 0.170  & 0.094 & 0.202 & 0.167 & 0.310  & 0.149 & 0.285 \\
			\textbf{AllSetTransformer} & 0.115 & 0.270  & 0.098 & 0.209 & 0.068 & 0.145 & 0.178  & 0.257 & 0.038 & 0.069 & 0.034 & 0.042 & 0.105 & 0.204 & 0.052 & 0.078 \\
			\cmidrule(l){1-17}
			\textbf{NS-HART (intra-edge)} & 0.073 & 0.153 & 0.057 & 0.106 & 0.050  & 0.117 & 0.125 & 0.189 & 0.120  & 0.256 & 0.048 & 0.098 & 0.177 & 0.381 & 0.082 & 0.169 \\
		\multirow{2}[1]{*}{\textbf{NS-HART}} & \textbf{0.538} & \textbf{0.647} & \textbf{0.569} & \textbf{0.713} & \textbf{0.224} & \textbf{0.313} & \textbf{0.502} & \textbf{0.685} & \textbf{0.217} & \textbf{0.354} & \textbf{0.385} & \textbf{0.567} & \textbf{0.241} & \textbf{0.414} & \textbf{0.386} & \textbf{0.508} \\
			 & \textbf{±0.02} & \textbf{±0.01} & \textbf{±0.02} & \textbf{±0.03} & \textbf{±0.04} & \textbf{±0.05} & \textbf{±0.05} & \textbf{±0.06} & \textbf{±0.04} & \textbf{±0.05} & \textbf{±0.01} & \textbf{±0.03} & \textbf{±0.03} & \textbf{±0.04}& \textbf{±0.11} & \textbf{±0.13} \\
			\bottomrule
		\end{tabular}%
		\label{tab_main_nofeature}%
	}
\end{table*}%

\subsection{Performance Comparison (Q1)}
\paragraph{\textbf{Performance on TR-EF Tasks.}}
In TR-EF tasks, models need to learn from both entity features and structural graph features. We follow the same setting as \citep{ref16}, and report the results in Table \ref{tab_main_featurel}. From these results, We make the following observations: \textbf{(i)} N-ary subgraph reasoning-based methods (such as UniSAGE, UniGAT, AllSetTransformer and NS-HART) generally show better performance than others, demonstrating their enhanced inductive inference ability. \textbf{(ii)} NS-HART achieves the best performance across all datasets. When n-ary facts make up a larger proportion, the performance gap becomes more pronounced. \textbf{(iii)} Triple-based methods generally show poor performance than the others. With an increasing number of n-ary relational facts, the performance gap between n-ary-based methods and triple-based methods generally becomes larger, corroborating the need for devising structures to handle n-ary facts. \textbf{(iv)} Graph neighborhood information is very helpful to the inductive link prediction task. By ablating the neighborhood information, NS-HART would degrade to NS-HART (intra-edge), causing a general performance decrease.

\paragraph{\textbf{Performance on TR-NEF Tasks.}}
In TR-NEF tasks, models need to capture and utilize n-ary structural patterns to make inferences. For this experiment, we set initial entity embeddings as the average outcomes of their linked semantic relations. Thus, predictions are based solely on relational information. As shown in Table \ref{tab_main_nofeature}, it can be observed that: \textbf{(i)} NS-HART consistently surpasses other methods, proving its ability to mine n-ary relational correlations. \textbf{(ii)} When entity features are ablated, the performance gap between subgraph reasoning-based methods and others becomes more pronounced. This highlights the importance of directly capturing multi-hop information via subgraph aggregating networks, particularly when graph structures become the main clues. \textbf{(iii)} In most datasets, NS-HART achieves even better performance than TR-EF, again proving its strong relational structure reasoning capabilities.

\begin{figure}\centering
	\includegraphics[width=0.7\linewidth]{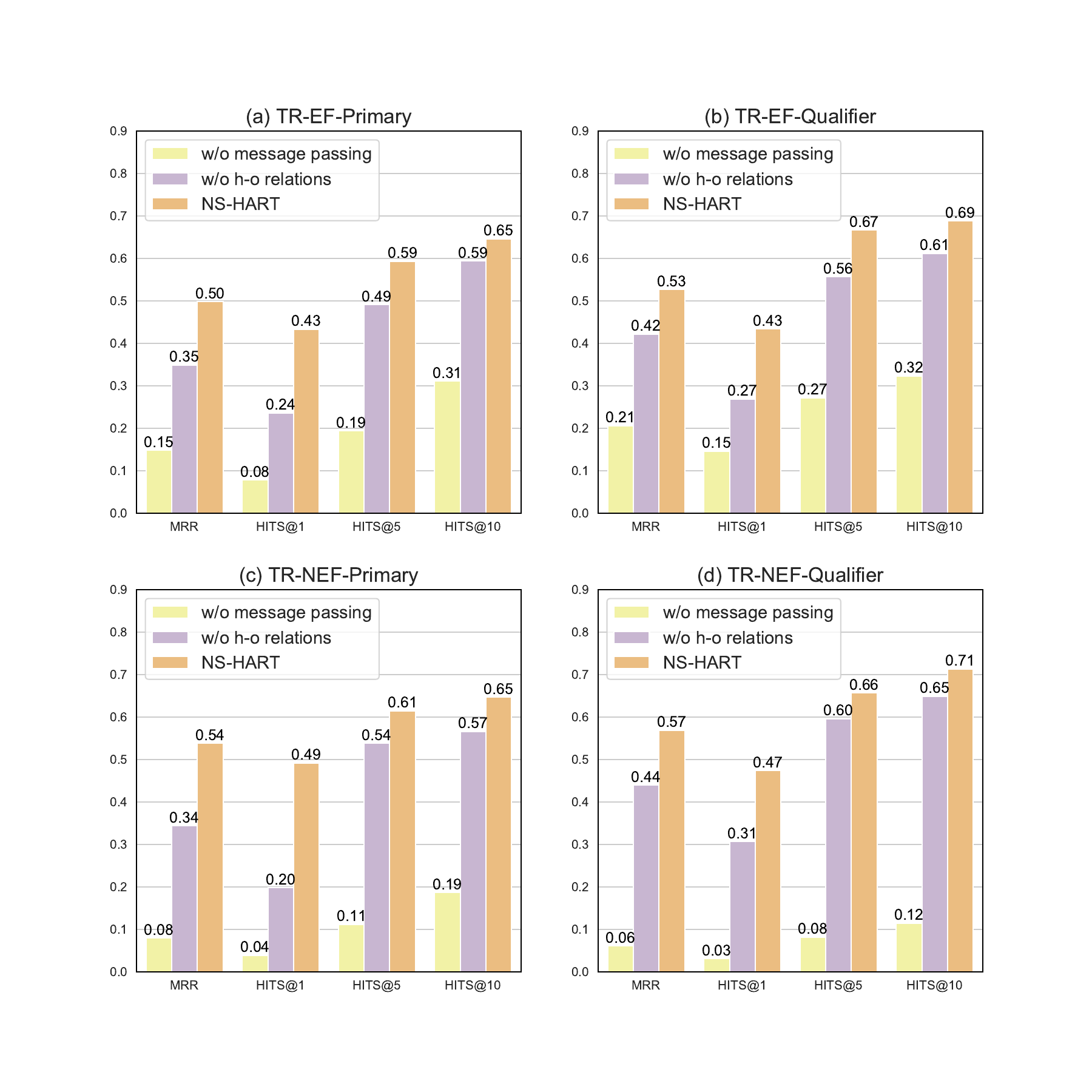}
	\caption{Results of the TR-EF and TR-NEF tasks on the WD20K (100) V1 using NS-HART, along with ablation studies for "w/o message passing" and "w/o high-order relations".}
	\label{fig_exp_ablation}
\end{figure}

\begin{table}
    \centering
    \caption{Results (AUC-PR) of PSR tasks. "P/Q" denotes results for primary or qualifier entity prediction.}
    {\fontsize{10pt}{12pt}\selectfont
    \resizebox{0.48\textwidth}{!}{
        \begin{tabular}{c|c|c|c|c}
            \toprule
            \multirow{3}{*}{\textbf{Method}} & \multicolumn{1}{c|}{\textbf{WD20K}} & \multicolumn{1}{c|}{\textbf{WD20K}} & \multicolumn{1}{c|}{\textbf{FI-MFB}} & \multicolumn{1}{c}{\textbf{FI-MFB}} \\
            & \multicolumn{1}{c|}{\textbf{(100) V1}} & \multicolumn{1}{c|}{\textbf{(66) V1}} & \multicolumn{1}{c|}{\textbf{(100)}} & \multicolumn{1}{c}{\textbf{(33)}} \\
            \cmidrule{2-5}
            & \textbf{P / Q} & \textbf{P / Q} & \textbf{P / Q} & \textbf{P / Q} \\
            \midrule
            \textbf{UniSAGE} & 60.1 / 73.5 & 62.8 / 68.4  &  60.1/ 68.2 & 64.3/ 60.2 \\ 
            \textbf{UniGAT} & 74.6 / 80.4 & 80.2 / 86.4  &  64.4/ 74.8 & 94.4 / 87.7 \\
            \textbf{AllSetTransformer} & \textbf{80.6} / 78.4 & 82.9 / 87.8 & 93.9 / 90.7  & 95.3 / 87.9 \\ 
            \textbf{NS-HART} & 80.1 / \textbf{87.3} & \textbf{92.4} / \textbf{95.1}  & \textbf{96.7} / \textbf{94.9}  & \textbf{98.7} / \textbf{97.4}  \\
            \bottomrule
        \end{tabular}
    }}
    \label{tab_SR}
\end{table}

\paragraph{\textbf{Performance on PSR Tasks.}}
These tasks focus on logical inference on local subgraphs of targeted links. As discussed in \textbf{Sec 5.2}, the expressive power of the aggregating multiset function is crucial. To validate this assumption, we compare NS-HART with other HGNNs and report the results in Table \ref{tab_SR}. The results corroborate our assumption: As the expressive power of the multiset functions increases (progressing from UniSage (Sum) to UniGAT (Attention) to AllSetTransformer (Set Transformer)), performance generally improves as well. Additionally, NS-HART consistently surpasses other methods, highlighting the superiority of employing the proposed Relational Transformer as the aggregating function.

\begin{table}
    \centering
    \caption{Performance of different positional encoding mechanisms on the WD20K (100) V1 for the TR-EF (MRR), TR-NEF (MRR), and PSR (AUC-PR) tasks.}
    {\fontsize{6pt}{8pt}\selectfont
    \resizebox{0.48\textwidth}{!}{
        \begin{tabular}{c|c|c|c}
            \toprule
            \multirow{2}{*}{\textbf{Method}} & \multicolumn{1}{c|}{\textbf{TR-EF}} & \multicolumn{1}{c|}{\textbf{TR-NEF}} & \multicolumn{1}{c}{\textbf{PSR}}\\
            \cmidrule{2-4}
            & \textbf{P / Q} & \textbf{P / Q} & \textbf{P / Q} \\
            \midrule
            \textbf{Random (NS-HART)} & \textbf{0.498} / \textbf{0.527} & 0.538 / 0.569  &\textbf{80.1} / \textbf{87.3} \\ 
            \textbf{Simple} & 0.244 / 0.324  & \textbf{0.543} /\textbf{ 0.57}4  & 79.3 / 80.9 \\
            \textbf{Same} & 0.102 / 0.095 & 0.529 / 0.561 & 78.6 / 84.0 \\   
            \bottomrule
        \end{tabular}
    }}
    \label{tab_PE}
\end{table}

\subsection{Ablation Study (Q2)}
As discussed in Sec. 4.2, NS-HART possesses two main advantages: the ability to preserve multi-hop hypergraph structures and to model interactions within each hyperedge. To demonstrate this assumption, we take TR-EF and TR-NEF tasks of WD20K (100) as an example and compare NS-HART with two variants: \textbf{(i) "w/o message passing"} treats all nodes in the sampled n-ary semantic subgraph and intra-edge relations as a sequence and employ a Transformer to get the output, resembling the graph Transformers \citep{ref46}. \textbf{(ii) "w/o high-order relations"} runs NS-HART on an adjusted semantic hypergraph, where each original hyperedge is split into several hyperedges containing binary relational facts. The remaining hyper-parameters are kept invariant. We present the outcomes in Figure \ref{fig_exp_ablation}. The results suggest that the message-passing mechanism and high-order relations all play significant roles. Ablating the message-passing mechanism results in significant performance degradation, highlighting the importance of capturing node-hyperedge associations. While NS-HART accounts for these factors, graph Transformers may require additional support.

To evaluate the impact of the role-aware positional encoding mechanism in NS-HART, we compare it with two variants: \textbf{(i) "Simple"} assigns all entities a positional mark of 0 and all relations a mark of 1. \textbf{(ii) "Same"}
assigns all entities and all relations a positional mark of 0. The results for various tasks on WD20K (100) are presented in Table \ref{tab_PE}. The findings reveal that the "Same" variant leads to significant performance drops across all datasets, while the "Simple" variant shows notable decreases in TR-EF and PSR tasks. In TR-NEF tasks, however, entities are initialized based on their linked relations, reducing the importance of role-aware mapping.

\begin{figure}\centering
\includegraphics[width=0.95\linewidth]{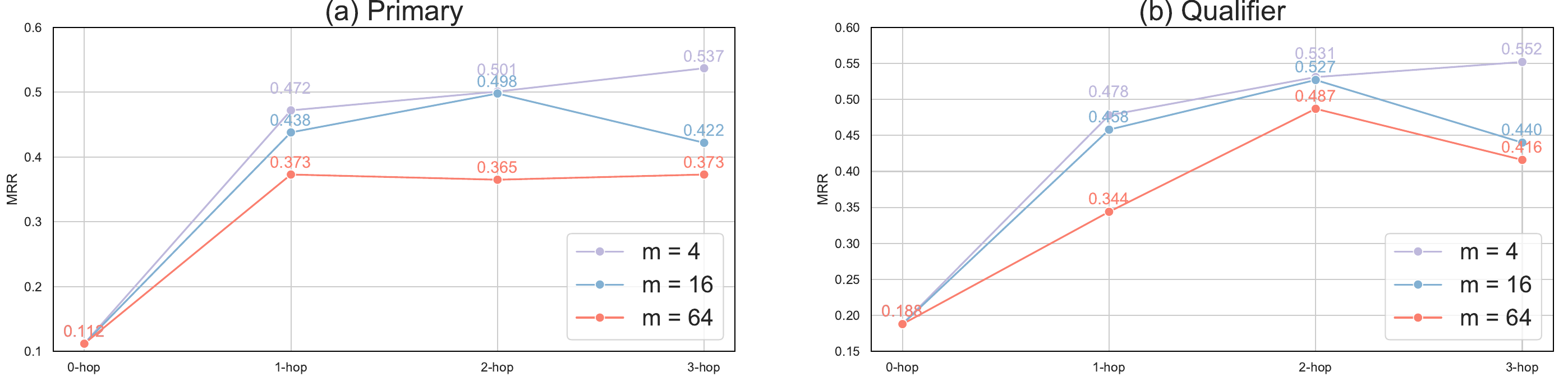}
	\caption{Performance (on WD20K (100) V1 of TR-EF tasks) of NS-HART with varied hops under different sampling scales.}
	\label{fig_exp_hop}
\end{figure}

\begin{figure}\centering
\includegraphics[width=0.95\linewidth]{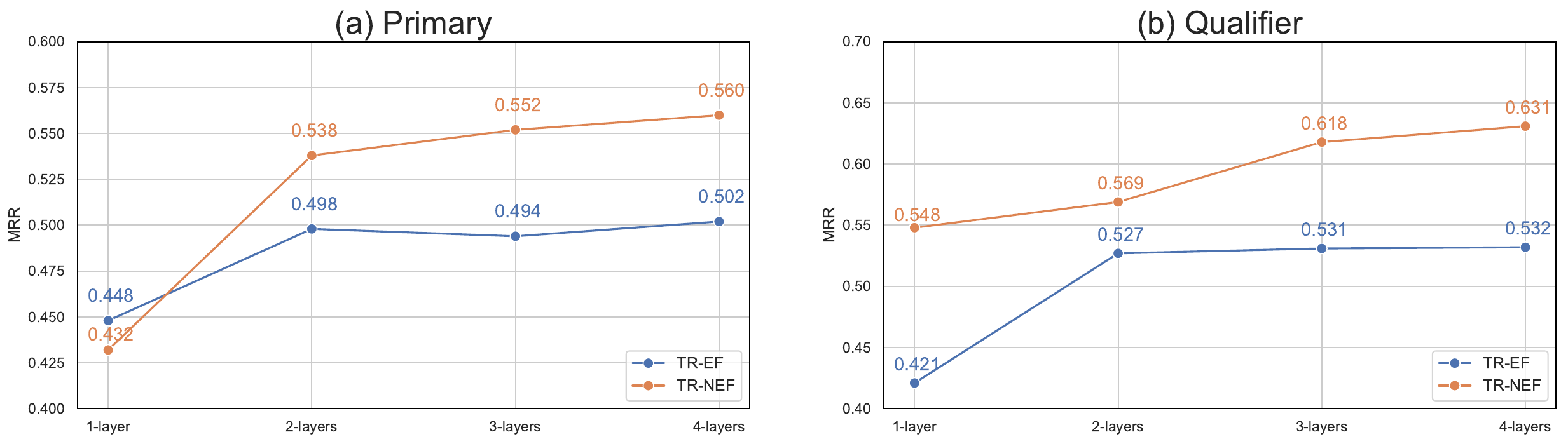}
	\caption{Performance (on WD20K (100) V1 of TR-EF tasks and TR-NEF tasks) of NS-HART with varied Transformer layer settings.}
	\label{fig_trans_layer}
\end{figure}

\subsection{Key Parameter Analysis (Q3)}
In this section, we study the effect of parameters like sampling scale and Transformer layers.

To investigate the impacts of the hop number and the sampling scale, we take TR-EF tasks on WD20K (100) V1 as an example. We conduct experiments with the hop number $k$ varying from 0 to 3, and the basic sampling number $m$ ranging in [4, 16, 64]. From the results of Figure \ref{fig_exp_hop}, we have the following observations: \textbf{(i)} When m=16 and 64, the performance generally first rises and then decreases. It indicates that neighborhood information is truly helpful, but information from overly distant neighbors can also introduce much noise for embeddings with limited length. \textbf{(ii)} The best results are obtained when m=4 and k=3, demonstrating that a sampling process is needed to prevent overfitting.

To study the effect of Transformer layers, we take TR-EF and TR-NEF tasks on WD20K (100) V1 as an example. As shown in Figure \ref{fig_trans_layer}, increasing the layers from 1-hop to 2-hop results in a significant performance improvement, demonstrating NS-HART's superiority over previous Transformer-like attention mechanisms in graphs \citep{ref61,ref35} (which use only one layer self-attention). However, further increasing the layers to 3 or 4 yields minimal performance gains, indicating NS-HART's efficiency with just two layers for ILP tasks.

\begin{figure}\centering
	\includegraphics[width=0.9\linewidth]{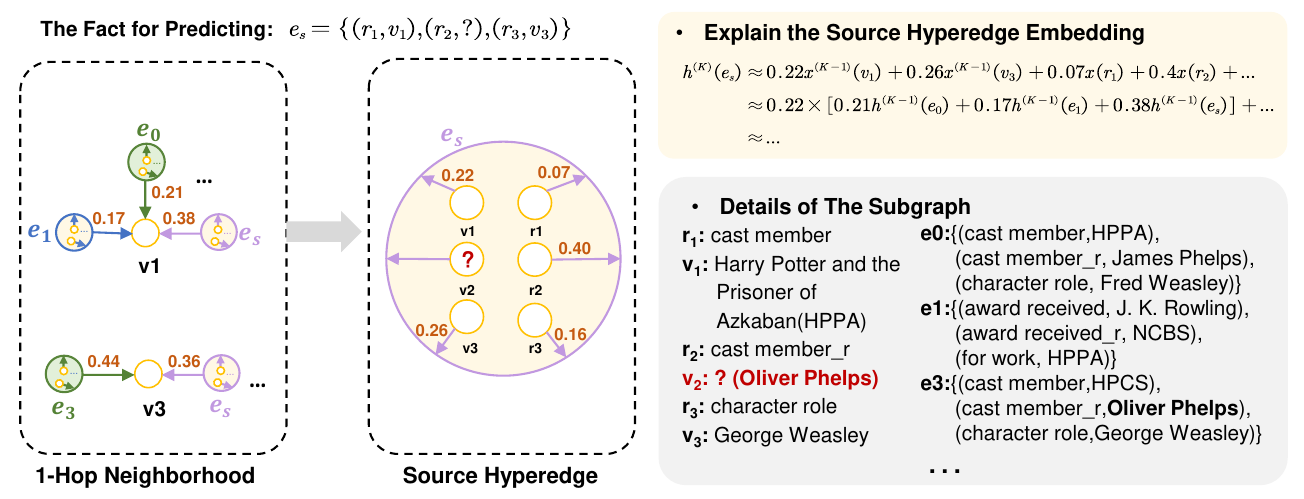}
	\caption{Case Study: An example to visualize the average attention scores of the Transformer.}
	\label{fig_exp_explain}
\end{figure}

\section{Case Study of NS-HART (Q4)}
\label{sec:case}
One advantage of NS-HART is the ability to aggregate multi-hop attention scores during message passing, which may aid users in understanding the results. If we consider the average values from the multi-head attention mechanism in Transformers as weights, the source hyperedge embedding $h_{e_s}$ can be approximately derived using nested attention scores. To detail the explanation process, we provide an example in Figure \ref{fig_exp_explain}. For visualization, we set the maximum hop to 1, the Transformer layer number to 1. Suppose the given query facts are:\{(cast\_member, HPPA),(cast member\_r, \textbf{?}), (character role, George Weasley)\}. We can observe that in the intra-edge neighborhood, the target entity’s semantic role "cast member\_r" contributes the most, aligning with human intuition. Additionally, a hyperedge related to entity "v3" contains the exact target entity "Oliver Phelps", which implies that" Oliver Phelps casts George Weasley in another movie HPCS". Through NS-HART’s multi-hop aggregation mechanism, the updated embedding of entity "v3" carries this vital clue.

\section{Conclusion and Discussion}
In this work, we delve into an underexplored challenging task, fully ILP on n-ary relational facts. Unlike transductive embedding-based methods, we propose an n-ary subgraph reasoning-based solution to capture entity-independent relational patterns. It includes two key innovations: the notion of n-ary semantic hypergraphs and n-ary subgraph aggregating networks, NS-HART. Theoretical analysis from the score function perspective and extensive experiments on three ILP sub-tasks demonstrate the superior inductive inference ability of subgraph reasoning using NS-HART. This work has several broader impacts: First, the concept of n-ary semantic hypergraphs offers new ideas for richer knowledge graph representations. Second, integrating Sequence Transformers into message-passing processes also holds the potential for further synergy between GNNs and LLMs in reasoning over semantic hypergraphs.

\begin{acks}
This work was supported by the Natural Science Foundation of Heilongjiang Province of China (Grant No. LH2023F018), and the Fundamental Research Funds for the Central Universities (Grant No. LH2023F018).
\end{acks}

\bibliographystyle{ACM-Reference-Format}
\bibliography{sample-base}

\appendix
\begin{table*}
	\caption{Statistics of the datasets. "H\%" denotes the proportion of n-ary facts. "avg" denotes the average arity of facts.}
	\large
	\label{tab_dataset}%
	\resizebox{\textwidth}{!}{
		\begin{tabular}{lccccccccccccc}
			\toprule
			\multirow{2}[1]{*}{\textbf{Splits}}& \multicolumn{4}{c}{\textbf{WD20K (100) V1}} & \multicolumn{4}{c}{\textbf{WD20K (66) V1}} & \multicolumn{4}{c}{\textbf{WD20K (100) V2}} \\ 
			\cmidrule(r){2-5} \cmidrule(l){6-9}  \cmidrule(l){10-13}
			& \textbf{All Facts (H\%)} & \textbf{Entities} & \textbf{Relations} & \textbf{Arity (avg)} & \textbf{All Facts(H\%)} & \textbf{Entities} & \textbf{Relations} & \textbf{Arity (avg)} & \textbf{All Facts(H\%)} & \textbf{Entities} & \textbf{Relations} & \textbf{Arity (avg)}\\
			\midrule
			\textbf{Train} & 7,785 (100\%)  & 5783  & 92  & \multirow{4}[1]{*}{3-7 (3.6)}  & 9,020 (85\%) & 6522  & 179  & \multirow{4}[1]{*}{2-7 (3.1)} & 4,146 (100\%)  & 3227  & 57 & \multirow{4}[1]{*}{3-7 (3.6)} \\
			\textbf{Inference} & 2,667 (100\%) & 4218  & 75  &   & 6,949 (49\%) & 8313  & 152 &  & 4,274 (100\%) & 5573  & 54 & \\
			\textbf{Validation} & 295 (100\%) & 643   & 43  &   & 910 (45\%) & 1516  & 111  &   & 538 (100\%) & 973   & 43 & \\
			\textbf{Test}  & 364 (100\%) & 775   & 43  &  & 1,113 (50\%) & 1796  & 110 &  & 678 (100\%) & 1212  & 42 & \\
			\midrule
			\midrule
			\multirow{2}[1]{*}{\textbf{Splits}}& \multicolumn{4}{c}{\textbf{WD20K (66) V2}} & \multicolumn{4}{c}{\textbf{FI-MFB (100)}} & \multicolumn{4}{c}{\textbf{FI-MFB (33)}} \\ 
			\cmidrule(r){2-5} \cmidrule(l){6-9}  \cmidrule(l){10-13}
			& \textbf{All Facts (H\%)} & \textbf{Entities} & \textbf{Relations} & \textbf{Arity (avg)} & \textbf{All Facts(H\%)} & \textbf{Entities} & \textbf{Relations} & \textbf{Arity (avg)} & \textbf{All Facts(H\%)} & \textbf{Entities} & \textbf{Relations} & \textbf{Arity (avg)}\\
			\midrule
			\textbf{Train} & 4,553 (65\%)  & 4269  & 148 & \multirow{4}[1]{*}{2-7 (2.9)}   & 9281 (100\%)  & 1983  & 33 & \multirow{4}[1]{*}{3-5 (3.0)}    & 15322 (26.49\%) & 4179  & 79 & \multirow{4}[1]{*}{2-4 (2.3)} \\
			\textbf{Inference} & 8,922 (58\%) & 9895  & 120  &   & 2909 (100\%) & 1007  & 34  &  & 4023 (37.86\%) & 1704  & 61 & \\
			\textbf{Validation} & 1,480 (66\%) & 2322  & 79 &   & 935 (100\%) & 597   & 22  &   & 1200(40.25\%) & 739   & 45 &  \\
			\textbf{Test}  & 1,840 (65\%) & 2700  & 89  &    & 1157 (100\%) & 631   & 26  &    & 1494 (39.63\%) & 790   & 49 &  \\
			\bottomrule
		\end{tabular}%
	}
\end{table*}

\section{Overall Training Process}
\label{sec:training}
\subsection{Training Regime}
In the training phase, for each fact, we generate multiple learning instances by ablating its entities at different positions. The training processes are based on the negative sampling strategy. The goal of the training process is to optimize \textbf{the relation embedding matrix, the parameters of subgraph aggregating networks, and the linear projection layer (when entity textual information is available)}. This is achieved by maximizing the scores of true targets and minimizing the scores of the false ones for each instance.  We employ a binary cross entropy loss as our training objective:
\begin{equation}
	\mathcal{L} = \sum_{j=1}^{|\hat{V}_s|}{y_j\log}\mathcal{P}_j\left( e_s,\hat{V}_s \right) 
     -\left( 1-y_j \right) \log \left[ 1-\mathcal{P}_j\left( e_s,\hat{V}_s \right) \right]
\end{equation}

Where $ \hat{V}_s $ is the sampled candidate entity set, and $j$ is one of a candidate entity. We assign  $y_j=1$  if  $ \left( r_p,v_j \right)$  combined with the remaining of $e_s$  forms a true fact, otherwise $y_j=0$. 

For TR tasks, we randomly select other entities from the KGs as negative examples. For PSR tasks, we randomly select entities from the input subgraphs (constructed from the neighborhoods of query entities and the positive targets) as negative examples.

Note that, in the actual training phase, we adopt the mini-batch training strategy \citep{ref41}. For PSR tasks, we treat each subgraph in the batch independently, combining them for the input. For TR tasks, we integrate a mini-batch of source hyperedges into the known base graph and merge all subgraphs sampled from each source hyperedge as the input subgraph. In contrast, the evaluating phase is single fact-based for TR tasks as mini-batch testing may inadvertently bring additional information from the test data.

\subsection{Initialization}
In all tasks, the relation embeddings are initiated using Glorot initialization. For TR-EF, we initialize the entity embeddings by projecting their original features to the relation space. For TR-NEF, we initiate each entity embedding by averaging its linked semantic roles. For PSR, features of each node $u$ in the subgraph are initialized using the hash map values of $d(u, e_s)$ following previous work \citep{ref19}. Here, $d(u, e_s)$ represents the hop counts between node $u$ and the source hyperedge $e_s$. In future work, more advanced inductive initialization methods, such as the query-dependent approaches described in HCNet \citep{ref54}, can be explored.

\section{Dataset Details}
\label{sec:dataset}
As shown in Table \ref{tab_dataset}, each of these datasets consists of a training set, an inference set (with entities entirely distinct from the training set), a validation set, and a test set (with entities already seen in the inference set). Following previous work \citep{ref16}, the construction of each dataset proceeds as: 
\textbf{(i) Training Set Formation.} Filter out a certain quantity of facts from the original datasets according to the desired n-ary proportion. Then, sample a subset of entities and their $l_{tr}$-hop neighborhood to build the training set. 
\textbf{(ii) Inference Part Formation.} Filter out entities in the training set, and sample another subset of entities with their $l_{inf}$-hop neighborhood for the inference part. 
\textbf{(iii) Inference Part Division.} Split facts in the inference part with a ratio of about 55\%/20\%/25\% into inductive inference, validation, and test sets, respectively. Note that, the main triplets of n-ary facts do not overlap between the inference set and the validation/test sets.

\section{Experimental Details}
\label{sec:experiment}
\subsection{Details of Baseline Methods}
\label{sec:baseline}
In the main experiments, we aim to study the benefits of employing n-ary semantic hypergraph, and the effects of NS-HART in inductive learning on this graph. Therefore, we first choose previous triple-based methods and hyper-relational-based methods as baselines. To learn on the proposed structure, we also consider some popular spatial-based HGNNs. To handle semantic relations, we modify the original HGNNs by treating both entities and relations within a hyperedge as vertices during the $\mathcal{V}\rightarrow \mathcal{E}$ process. Though there are more recent methods on transductive settings (e.g. HAHE \citep{ref13} and HyperFormer \citep{ref14}), they are not selected for our study as their enhancements are not tailored for inductive scenarios. The details of baseline methods are listed as follows:
\begin{itemize}
	\item \textbf{BLP} \citep{ref38}: The simplest triple-based method, which employs a linear projection for each entity and calculates plausibility scores for each triple using transition-based functions. 
	\item \textbf{CompGCN} \citep{ref39}: A triple-based method that first employs GCN for multi-relational graphs to derive entity and relation embeddings with structural information, and then adopts a decoder to calculate plausibility scores. It can be seen as the triple-version of StarE.
	\item \textbf{QBLP} \citep{ref16}: A hyper-relational-based method that combines the linear projection and the Transformer to handle n-ary facts, also known as qualifier-aware BLP.
	\item \textbf{StarE} \citep{ref6}: A hyper-relational-based method using the GCN encoder + decoder framework similar to CompGCN. For the GCN encoder, it introduces specially designed GCN for hyper-relational KGs. For the decoder part, it employs the Transformer for n-ary facts. 
	\item \textbf{GRAN} \citep{ref11}: A hyper-relational-based method that incorporates Transformer decoders with more sophisticated edge-biased self-attention mechanisms compared to QBLP.
	\item \textbf{UniSage} \citep{ref22}: A spatial-based HGNN which generalizes GraphSage to hypergraphs.
	\item \textbf{UniGAT} \citep{ref22}: A spatial-based HGNN which generalizes GAT to hypergraphs. It also adopts the attention mechanism to assign importance scores during aggregation processes.
	\item \textbf{AllSetTransformer} \citep{ref31}: A spatial-based HGNN which uses Set Transformer (with only one self-attention layer and one query vector) as a learnable multiset function.
\end{itemize}

\subsection{Parameter Settings}
\label{sec:param}
We fix the following configurations for all models across all the datasets: the batch size is set to 128, the embedding size is set to 200, the maximum arity number is set to 7, the Transformer layer number $L$ is set to 2, the attention head number is set to 4, and the Transformer hidden size is set to 512. When implementing the Transformer, we use FlashAttention \citep{ref40}, a parallelism and work partitioning approach. The maximum epoch is set to 300 for TR-EF/TR-NEF and 50 for PSR. In the main experiments, we set the max hop number $K=2$ and set the basic sampling number $m=16$ . The max neighboring hyperedges in $\mathcal{E}\rightarrow \mathcal{V}$ process are set to $2m$. For optimization, we employ the negative sampling strategy with 50 (for TR-EF/TR-NEF) or 1 (for PSR) false entities for each fact, and adopt AdamW \citep{ref42} combined with the plateau strategy. On each dataset, we select the following parameters in their respective ranges using the grid search method for each method, i.e. the learning rate $lr$: [1e-5, 5e-5, 1e-4, 5e-4], the dropout rate $\rho$: [0.1, 0.2, 0.3]. We mainly use a Nvidia 4090 GPU (24G) or L20 GPU (48G) to train the models and hyperparameters are tuned on the validation set according to the best MRR (for TR-EF/TR-NEF) or AUC-PR (for PSR).

\section{Discussions on The N-ary Semantic Hypergraph}
\label{sec:NS-Graph}
The n-ary semantic hypergraph diverges from hyper-relational KGs and relational hypergraphs, offering several distinct advantages for fully inductive scenarios:
\textbf{(i) Facilitate neighborhood expansion.} It allows for the straightforward expansion of an entity's multi-hop neighborhood within the semantic hypergraph, a critical factor for inductive LP tasks \citep{ref19}. This contrasts with hyper-relational KGs, which treat qualifier entities as auxiliary edge information, thereby neglecting qualifier entities’ neighborhoods that may hold critical information.
\textbf{(ii) More expression flexibility.} Compared with hyper-relational KGs, the n-ary semantic hypergraph does not necessitate identifying a main triplet, proving more robust in scenarios where primary facts cannot be identified, such as “A, B, and C cooperate in project AIP”. Compared with relational hypergraphs, the n-ary semantic hypergraph contains the intra-role of each entity within a fact, which are more expressive. This characteristic also paves the way for richer KG representations, and we leave more applications of the n-ary semantic hypergraph for future work.
\textbf{(iii) Synergy with HGNNs.} The n-ary semantic hypergraph retains the core structure of traditional hypergraphs, enhancing synergy with HGNNs and enabling us to solve inductive LP tasks from hypergraph learning. In contrast, hyper-relational KGs remain fundamentally triple-based with binary edges.

\section{Deeper Analysis of NS-HART}
\label{sec:NS-HART}

\subsection{NS-HART vs. Graph Transformers}
\label{sec:vs_GT}
Recently, a line of work has begun employing Transformers to handle graph data, namely Graph Transformers \citep{ref46,ref48}. Diverging from message-passing neural networks (MPNNs) in GNNs, they employ Transformers to capture pairwise relationships among nodes while incorporating structural encodings as biases to account for graph structure. While effective for many graph tasks, they struggle to represent complex hyperedge information. For example, NodePiece \citep{ref58} employs a Graph Transformer-like architecture for KGs, treating entities and relations in the neighborhood as input tokens. However, if applied to n-ary semantic hypergraph, this approach loses crucial information about which nodes and relations belong to each hyperedge. In contrast, NS-HART retains the message-passing framework. By using Transformers as aggregation functions of MPNNs, NS-HART combines the strengths of both paradigms: Transformers for modeling interactions within a set, and MPNNs for multi-hop interactions across the graph.

\subsection{Efficiency}
\label{sec:time}
The efficiency of NS-HART depends heavily on the Transformer used in its message-passing process. Consider a sampled subgraph with $M$ hyperedges, $N$ entities, an average of $\bar{n}$ entities per hyperedge, and an average of $\bar{m}$ hyperedges connected to each entity. Over K iterations of $\mathcal{V}\rightarrow \mathcal{E}$ and the $\mathcal{E}\rightarrow \mathcal{V}$ processes, the computational complexities are roughly $O(KLd^2×(M\bar{n}^2+N\bar{m}^2))$. In comparison, UniSAGE has a complexity of $O(KLd×(M\bar{n}+N\bar{m}))$, and graph Transformers scale with $O(Ld^2×N^2)$. Without optimization strategies, NS-HART's scalability is limited, especially with dense graphs where $M$ and $\bar{m}$ are large. As a result, it is currently more suitable for subgraph-based tasks. For ILP tasks, regardless of the graph's size, most target entities are within the K-hop neighborhood, enabling applying NS-HART without significant additional computational costs. In future work, more efficient Transformer attention mechanisms can be incorporated to enhance NS-HART’s scalability for larger graphs and more tasks.

\end{document}